# Early Detection of Branched Broomrape *(Phelipanche ramosa)* Infestation in Tomato Crops by Using Leaf Spectral Analysis and Machine Learning

Mohammadreza Narimani*, Alireza Pourreza*, Ali Moghimi*, Parastoo Farajpoor*, Hamid Jafarbiglu*, Mohsen B. Mesgaran**

*Department of Biological and Agricultural Engineering, University of California, Davis, CA, 95616, United States ({mnarimani, apourreza, amoghimi, pfarajpoor, jafarbiglu}@ucdavis.edu)
**Department of Plant Sciences, University of California, Davis, CA, 95616, United States (mbmesgaran@ucdavis.edu)

**Abstract**: This study explores the use of leaf-level spectral reflectance and ensemble machine learning to detect early-stage infestations of branched broomrape (*Phelipanche ramosa*) in tomato (Solanum lycopersicum) crops. A field experiment was conducted in Woodland, California, monitoring 300 tomato plants across multiple growth stages defined by Growing Degree Days (GDD). Leaf reflectance (400–2500 nm) was measured with a portable spectrometer and preprocessed using noise removal, interpolation, Savitzky-Golay smoothing, and correlation thresholding. Notable spectral differences appeared in the 1500 nm and 2000 nm water absorption bands, indicating reduced water content in infected leaves at early stages. An ensemble model integrating Random Forest, Extreme Gradient Boosting, Support Vector Machine (RBF kernel), and Naive Bayes achieved 89% accuracy at 585 GDD, with recalls of 86% (infected) and 93% (non-infected). Accuracy declined at later stages, reaching 69% with reduced recall (50%) for infected samples, likely due to spectral noise from senescence and weed interference. Despite challenges like data imbalance and environmental confounders, the study demonstrates the potential of proximal sensing and ensemble learning for timely broomrape detection, enabling early intervention and minimizing yield loss.

## 1. INTRODUCTION

Tomato (Solanum lycopersicum) is a major global horticultural crop, yielding about 189.1 million tons annually across 5 million hectares (Singh et al., 2025). Its broad use—from fresh produce to processed goods like sauces and juices—drives significant market value (Arah et al., 2015), with average per capita consumption of 20 kg globally (Dorais et al., 2008). In the U.S., tomatoes are highly profitable, with California producing nearly 90% of the national supply, or about 12 million metric tons annually (Agricultural Statistics Service, 2020; Newsom, 2022), and accounting for 96% of domestic and 30% of global processing output (Winans et al., 2020). U.S. consumption averages 31 kg per capita (Alaimo et al., 2008). However, the 2014 Sustainable Groundwater Management Act (SGMA) imposed stricter water use policies (Moran & Wendell, 2014), contributing to a production decline. As shown in Figure 1, U.S. output rose from 4 million tons in 1961 to nearly 16 million in 2014, then declined, highlighting the need for improved crop management and control of threats like diseases, weeds, and parasitic plants.

Despite their economic value, tomato crops face threats from various pathogens and parasitic organisms. Fungal diseases like Fusarium oxysporum and Alternaria solani can reduce yields, while the parasitic weed branched broomrape (*Phelipanche ramosa* (L.) Pomel) poses a major challenge to tomato production (Adewale Osipitan et al., 2021; Chaerani & Voorrips, 2006; Nowicki et al., 2012; Parker, 2012). Lacking chlorophyll, broomrape remains underground for most of its life cycle, attaching to tomato roots and extracting nutrients, often undetected until severe yield losses occur—up to 90% (Joel et al., 2007; Parker, 2009). Remote sensing technologies are increasingly used to enhance pest and disease management by capturing subtle physiological changes via reflectance, temperature, and other indicators (Mulla, 2013; Pinter et al., 2003; Xue & Su, 2017). Their capabilities have been significantly improved through AI tools like machine learning and deep learning, which enable early detection through pattern recognition in large datasets (Kamilaris & Prenafeta-Boldú, 2018).

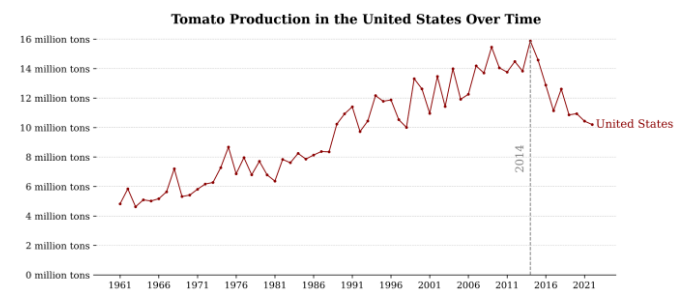

**Figure 1:** Historical trend of the United States tomato production (Food and Agriculture Organization of the United Nations, 2025).

Chemical control of broomrape shows some efficacy but is limited under certain conditions and may harm non-target species (Parker, 2009). Alternative approaches—such as crop rotation or resistant varieties—have produced mixed outcomes (Eizenberg & Goldwasser, 2018). In sunflower, hyperspectral imaging has successfully identified Orobanche cumana stress (Atsmon et al., 2022). Similarly, proximal leaf-level sensing offers clean reflectance signals without atmospheric interference (Lodovica et al., 2014), making it promising for early broomrape detection.

Because nearly two-thirds of broomrape's life cycle is underground and emergence is often hidden under the canopy, standard detection methods (e.g., RGB drones) are ineffective. To address this, the present study applies leaf-level spectral analysis and machine learning to identify broomrape before canopy symptoms appear. By analyzing reflectance differences between infected and non-infected plants across growth stages, we aim to pinpoint key spectral features and develop robust classification models for early detection—ultimately improving broomrape management in tomato crops.

## 2. METHODOLOGY

*2.1 Study Area and Data Collection*

The study was conducted on a tomato farm in Woodland, California, known for branched broomrape infestation (Figure 2). Growing Degree Days (GDD) were used to track tomato growth stages (Narimani et al., 2024), calculated daily using Equation (1):

$$GDD = \sum (\overline{T_i} - T_b) \quad (1)$$

Where $\overline{T_i}$ is the daily mean temperature (i refers to a given day) with $T_b$ representing base temperature of 10°C for tomatoes.

On May 21, 2023, we transplanted seedlings and randomly flagged 300 plants. At four key stages—585 GDD (vegetative), 897 GDD (flowering), 1216 GDD (fruit development), and 1568 GDD (ripening)—we collected two fully expanded leaves from the middle canopy of each plant, totaling 600 samples per stage (2400 overall). Leaves were cut at the stem (not nodes) to minimize water loss, placed in ice-cooled bags, and transported to the lab.

We used an HR-1024i full-range field-portable spectroradiometer (SVC Spectra Vista Corporation, NY, USA) covering 350–2500 nm. The system included an LC-RP PRO Leaf Clip, tungsten halogen illumination, a reversible white reference panel, and a fiber optic probe with a spot size of ~6.2 mm × 3.5 mm. All plants were tracked through harvest to determine broomrape infection status.

*2.2 Addressing Data Imbalance*

By the end of the season, 49 out of 300 monitored plants were confirmed as branched broomrape infected. To create a balanced dataset, we selected 49 non-infected plants whose mean reflectance fell within one standard deviation of the overall non-infected mean, ensuring a representative comparison group. This yielded 98 leaf samples per class for each GDD stage (196 total), providing sufficient data to train ensemble models effectively across all four growth stages.

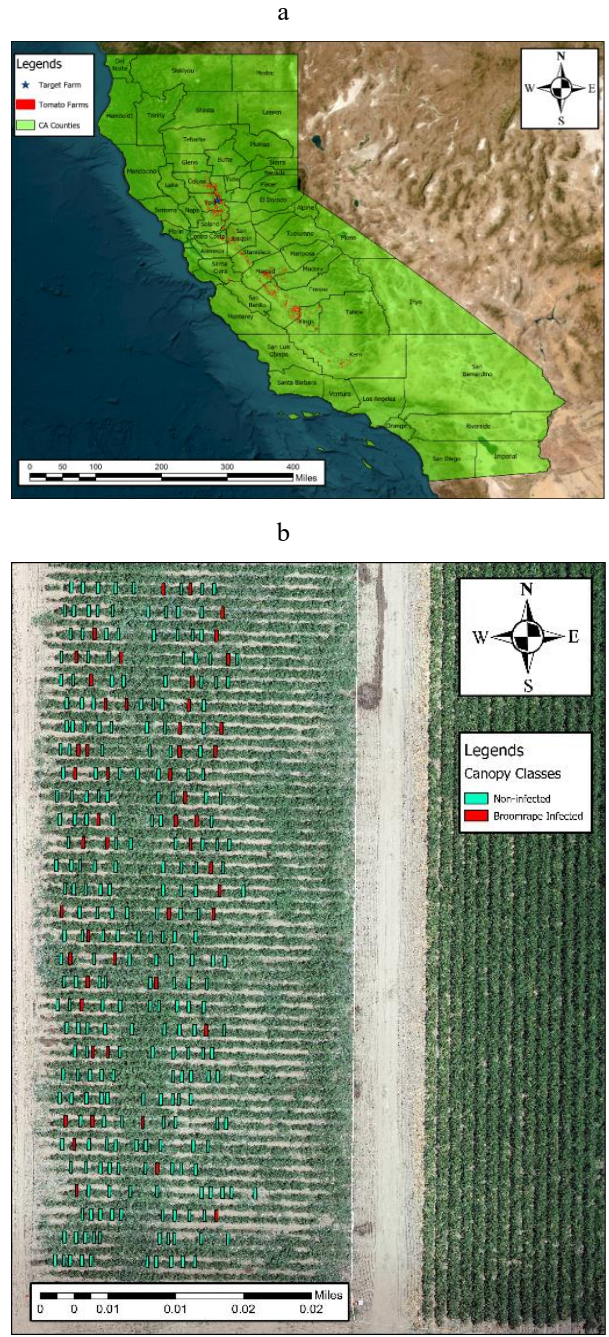

*Figure 2:* (a) Map of California showing major tomato farming counties and the location of our target farm, generated with ArcGIS Pro; (b) Top view of the target tomato farm, including selected non-infected and branched broomrape infected tomato plants.

*2.3 Spectral Data Preprocessing*

The spectroradiometer used in this study includes three base detectors, each covering a specific spectral range with distinct full-width half-max resolutions: approximately 3.3 nm at 700 nm, 9.5 nm at 1500 nm, and 6.5 nm at 2100 nm. Nominal

bandwidths are 1.5 nm (350–1000 nm), 3.8 nm (1000–1890 nm), and 2.5 nm (1890–2500 nm). To ensure consistency, we removed noisy bands (typically five before and after each detector range) and interpolated the remaining data to a uniform 1 nm resolution. We applied a Savitzky-Golay filter (quadratic, frame length = 7) to smooth the spectra while preserving important features, followed by standard scaling (StandardScaler) to normalize input features. To reduce redundancy, we implemented correlation thresholding: adjacent bands with >99% correlation were averaged. The correlation was computed using the Pearson method (Cohen et al., 2009). Figure 3 presents heatmaps of spectral band correlations at four GDD stages. For each stage, the numeric labels (e.g., 2100 → 106) show the dimensionality reduction—from the original number of bands to the final count after correlation filtering.

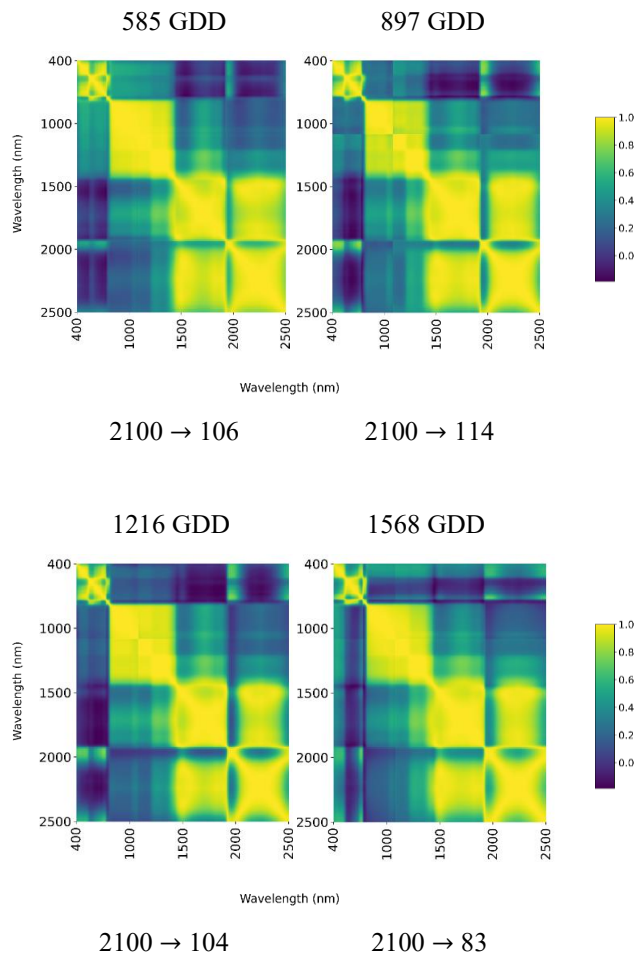

*Figure 3.* Correlation heatmaps of spectral bands at four key growth stages (585, 897, 1216, and 1568 GDD). Each heatmap shows pairwise correlations (400–2500 nm) from vegetative to ripening.

To quantify reflectance differences between non-infected and branched broomrape-infected leaves, we calculated the Relative Mean Difference using Equation (2). This metric was applied across the full wavelength range to isolate the most significant spectral differences:

$$RMD = \frac{\mu_{non} - \mu_{inf}}{\mu_{non}} \quad (2)$$

where RMD is the Relative Mean Difference, $\mu_{non}$ is the mean reflectance of non-infected leaves, and $\mu_{inf}$ is the mean reflectance of infected leaves.

*2.4 Machine Learning Model Development*

To classify infected and non-infected tomato leaves, we employed an Ensemble Learning Strategy that combines multiple diverse models using a voting-based meta-classifier. The dataset was split into 65% training, 15% validation, and 20% testing subsets. We began by evaluating 16 widely used base learners: Random Forest, XGBoost, Bagging Classifier, LDA, Decision Tree, LightGBM, SVM (RBF Kernel), Logistic Regression, Extra Trees, CatBoost, KNN, Naïve Bayes, Gradient Boosting, AdaBoost, Ridge Classifier, and a Neural Network. Each model was trained using an out-of-fold prediction strategy to prevent overfitting, allowing generalization to unseen data.

We computed average AUC scores and average pairwise correlations among model predictions to assess performance and redundancy. Models with low inter-correlation and high AUCs were favored, as their diversity enhances ensemble robustness. The best-performing and most complementary combination—Random Forest, XGBoost, SVM (RBF Kernel), and Naive Bayes—was selected. These were integrated using logistic regression as a meta-classifier to boost final accuracy (Dong et al., 2020).

3. RESULTS

The outcomes of our study are presented in two main categories: (1) a comparison of pure spectral reflectance between non-infected and branched broomrape-infected tomato leaves, and (2) an evaluation of the Ensemble Learning Strategy's performance in classifying infected and non-infected leaves.

Initially, we analyzed the full-range reflectance spectra to identify differences between non-infected and infected leaves. The relative mean differences between the two classes are shown in Figure 4. Notably, significant discrepancies emerged near the 1500 nm and 2000 nm wavelengths during the early growth stages—regions typically associated with water absorption. These findings suggest that branched broomrape-infected leaves exhibit reduced water content in the early stages, likely because the parasite draws additional water from the host plant (Mauromicale et al., 2008). As the plants advance to later growth stages, however, this trend reverses: non-infected plants display lower leaf water content as they channel resources toward fruit formation, whereas infected plants retain more leaf moisture due to diminished fruit development.

Following the spectral analysis, we applied the Ensemble Learning Strategy for leaf classification. After evaluating various base learners, the final selected models—Random Forest, XGBoost, Support Vector Machine (SVM) with a Radial Basis Function (RBF) kernel, and Naive Bayes, were chosen due to their minimal prediction correlation and structural diversity, which helped reduce the risk of overfitting.

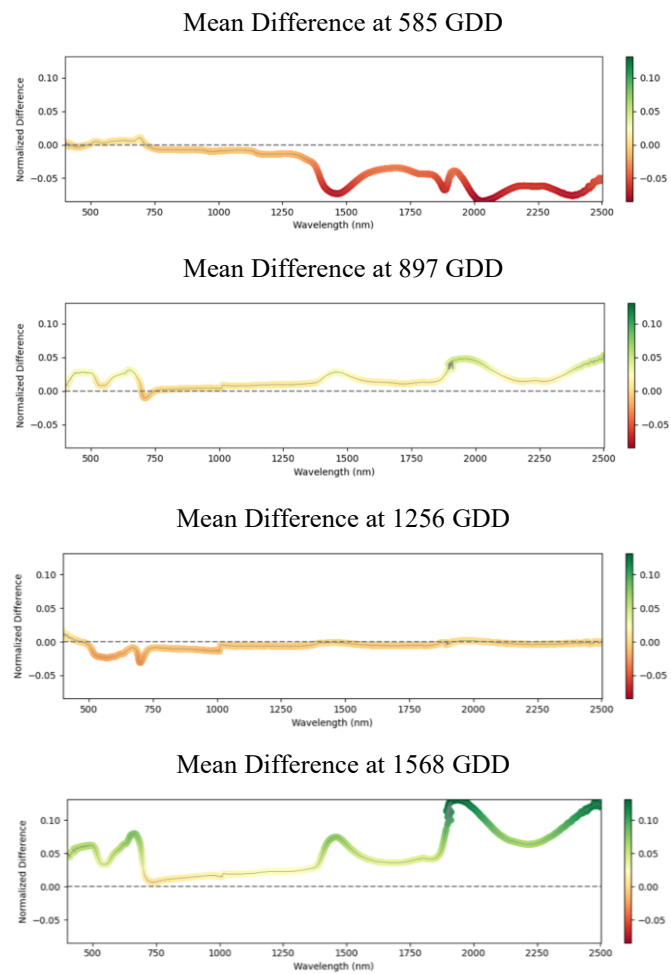

*Figure 4:* Relative mean differences in spectral reflectance between non-infected and branched broomrape infected tomato leaves.

By combining multiple algorithms with complementary strengths, the ensemble approach captures a broader range of data patterns and improves overall classification robustness. Additionally, the ensemble strategy enhances the model's ability to generalize to new, unseen data, which is critical for practical field applications. The feature importance analysis for these models, presented in Figure 5, underscores the pivotal role of water absorption regions in the classification process, particularly at crucial Growing Degree Day (GDD) stages. These regions consistently contributed to model performance across growth phases, reinforcing their reliability as discriminative features and highlighting their persistent relevance in phenological assessments.

Next, we assessed the model's predictive performance using the confusion matrix on a held-out test set to evaluate its accuracy and reliability.

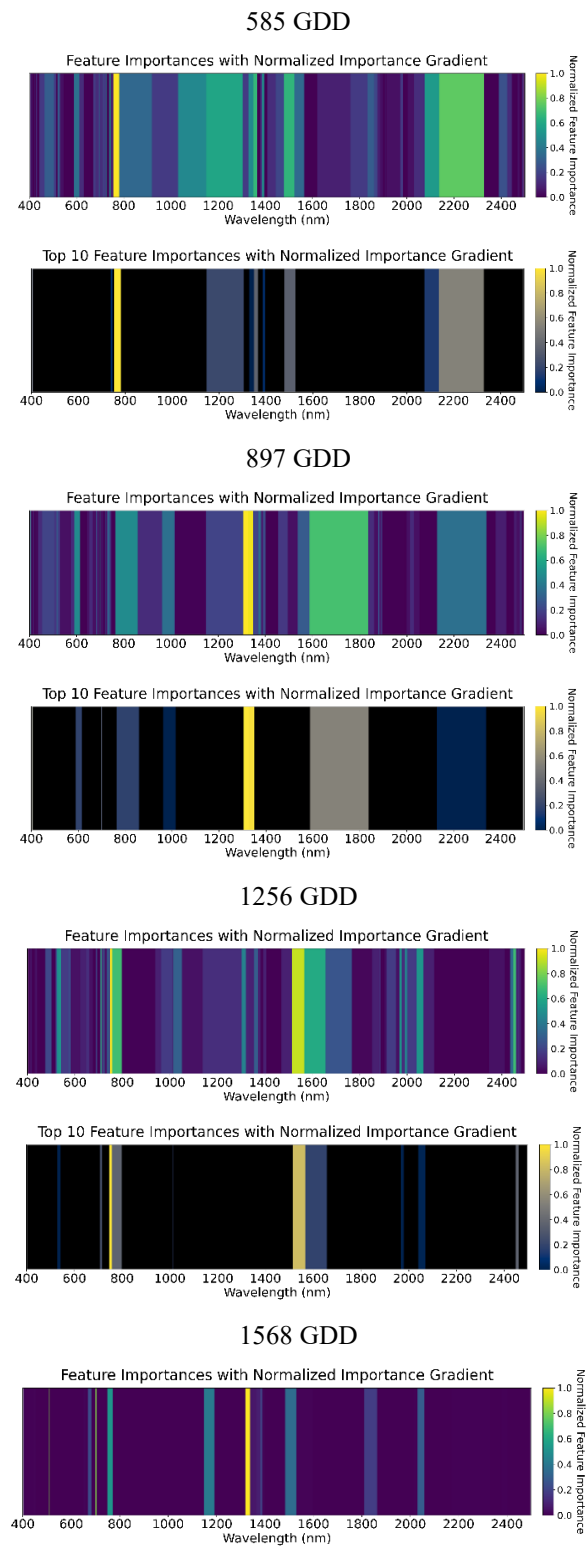

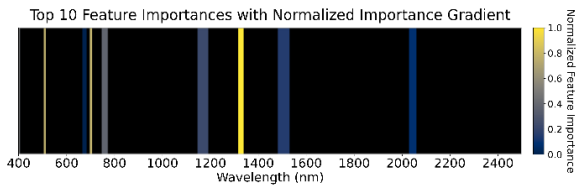

*Figure 5:* Feature importance analysis of final machine learning models across four key GDD stages.

At 585 GDD—our earliest leaf-level measurement—the models achieved an overall accuracy of 89%, with a recall of 86% for the branched broomrape-infected class (our target class) and a specificity of 93% for the non-infected class. This performance at an early growth stage is particularly encouraging for timely detection, which is our primary objective. However, model performance declined in later GDD stages, likely due to increased visual noise from other weeds and dried leaves. Figure 6 illustrates the confusion matrix across the four GDD stages, emphasizing the strength of early-stage detection.

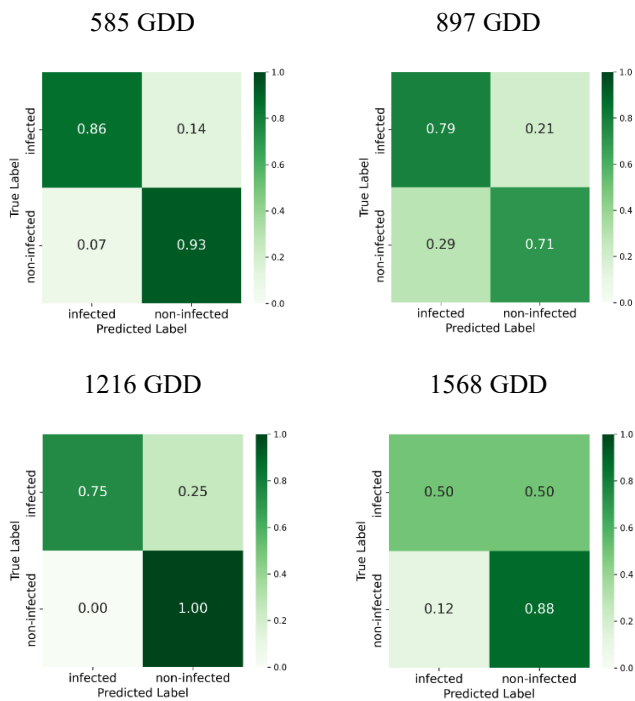

*Figure 6:* Confusion matrix of model performance across four key GDD stages.

## 4. DISCUSSION

This study confirms that leaf-level spectral data combined with machine learning effectively detects branched broomrape (*Phelipanche ramosa*) in tomato crops. Notably, significant differences were observed in the 1500 nm and 2000 nm water absorption bands during early growth stages, indicating lower water content in infected plants (Figure 4). These results align with prior findings that broomrape disrupts the host's water balance by siphoning off moisture and nutrients (Joel et al., 2007; Parker, 2009). Interestingly, as the crop matures, this trend reverses—infected leaves retain more moisture due to reduced allocation toward fruit development, suggesting a dynamic physiological response in host-parasite interactions.

Our Ensemble Learning Strategy—using Random Forest, XGBoost, SVM (RBF Kernel), and Naive Bayes—demonstrated strong classification performance, particularly at 585 GDD, with 89% accuracy and high recall values for both infected (86%) and non-infected (93%) classes. These models were selected based on their high AUC scores and low prediction correlation, enhancing generalizability and robustness. The most informative spectral features corresponded with water absorption wavelengths, further validating the biological basis of the spectral differences observed. While early-stage detection showed promising results, performance declined in later stages due to factors like weed interference and leaf senescence. Addressing these challenges could involve integrating additional sensing modalities, such as thermal or canopy-level hyperspectral data, and refining preprocessing steps. Data imbalance—stemming from relatively few infected samples—also posed a challenge. Although partially mitigated by careful sampling, future work may benefit from advanced augmentation techniques like synthetic oversampling. These findings support existing literature (Atsmon et al., 2022; Nowicki et al., 2012) showing that early, sensor-based detection of crop pathogens is both practical and impactful when tailored to specific plant-pathogen dynamics.

## 5. CONCLUSION

This study demonstrates that leaf-level reflectance analysis combined with an Ensemble Learning Strategy is effective for early detection of branched broomrape (*Phelipanche ramosa*) in tomato crops. Focusing on water absorption wavelengths, we identified distinct spectral signatures that differentiate infected from non-infected leaves before visible symptoms emerge. The ensemble of Random Forest, XGBoost, SVM (RBF Kernel), and Naive Bayes achieved high accuracy, particularly at 585 GDD, validating the potential of this multi-model approach.

This method offers a promising pathway for improving crop management and reducing broomrape-related losses. Early detection enables precise interventions and minimizes the environmental footprint of herbicide use. Nonetheless, issues such as class imbalance and confounding spectral noise at later stages highlight the need for further refinement. Future efforts should explore integrating other data sources (e.g., thermal or canopy-level imagery) and advanced models to better address complex field conditions. Overall, this approach holds strong potential for broader application in sustainable management of parasitic weeds across high-value crops.